\documentclass[letterpaper, 10 pt, conference]{ieeeconf}

\IEEEoverridecommandlockouts
\usepackage{amsmath,amssymb}
\usepackage{booktabs}
\usepackage{multirow}
\usepackage{graphicx}
\usepackage{balance}
\usepackage{placeins}
\usepackage{microtype}
\usepackage{xcolor}
\usepackage{tikz}
\usetikzlibrary{arrows.meta,positioning}
\usepackage[hidelinks]{hyperref}
\usepackage{url}
\usepackage{etoolbox}

\makeatletter
\patchcmd{\@maketitle}{\vskip0.25in}{\vskip0.10in}{}{\PackageError{riwanav}{Title spacing patch failed}{}}
\patchcmd{\@maketitle}{\vskip1.0em}{\vskip0.6em}{}{\PackageError{riwanav}{Author spacing patch failed}{}}
\makeatother
\IEEEaftertitletext{\vspace{-10pt}}

\renewcommand{\footnoterule}{%
  \kern-3pt
  \hrule width 0.6\columnwidth height 0.4pt
  \kern2.6pt
}

\hypersetup{
  pdftitle={RIWANav: Recursive World-Action Models with Self-Improvement for Urban Navigation},
  pdfauthor={Jing Xie, Shouwei Ruan, Yubin Wang, Yuxiang Zhang, Haitao Yang, Songchang Jin, Dianxi Shi},
  pdfsubject={Preprint}
}
\makeatletter
\def\ps@preprint{%
  \def\@oddhead{\small Preprint\hfil}%
  \let\@evenhead\@oddhead
  \def\@oddfoot{\hfil\small\thepage\hfil}%
  \let\@evenfoot\@oddfoot
}
\makeatother

\newcommand{\method}{\textsc{RIWANav}}
\newcommand{\dataset}{\textsc{UrbanNav}}
\newcommand{\wm}{\textsc{Cosmos}}
\newcommand{\meanstd}[2]{\ensuremath{#1 \pm #2}}
\newcommand{\bestmeanstd}[2]{\ensuremath{\mathbf{#1 \pm #2}}}
\newcommand{\secondmeanstd}[2]{\ensuremath{\underline{#1 \pm #2}}}

\newcommand{\tabvcenter}[1]{\raisebox{.65\ht\strutbox}[0pt][0pt]{#1}}
\newcommand{\R}{\mathbb{R}}

\title{RIWANav: Recursive World--Action Models with Self-Improvement \\
for Urban Navigation}

\author{\parbox{\textwidth}{\centering
Jing Xie$^{1}$, Shouwei Ruan$^{2}$, Yubin Wang$^{3,\ddagger}$, Yuxiang Zhang$^{4}$\\[1pt]
Haitao Yang$^{5}$, Songchang Jin$^{6}$, Dianxi Shi$^{6,*}$\\[4pt]
{\fontsize{9}{10.5}\selectfont
$^{1}$MoE Key Lab of Artificial Intelligence, Institute of AI, School of Computer Science,\\
Shanghai Jiao Tong University, Shanghai, China\\
$^{2}$Beihang University, Beijing, China\quad
$^{3}$The Hong Kong University of Science and Technology, Hong Kong, China\\
$^{4}$Tsinghua University, Beijing, China\quad
$^{5}$The University of Texas at Austin, Austin, USA\\
$^{6}$Intelligent Game and Decision Lab (IGDL), Beijing, China
}}\thanks{$^{\ddagger}$Project leader.\quad
$^{*}$Corresponding author.}}

\begin{document}
\pagestyle{preprint}
\maketitle
\thispagestyle{preprint}

\begin{abstract}
Long-horizon urban navigation requires sequential local decisions whose errors can compound over time. Imitation learning (IL) rarely learns from failures, while physical trial-and-error reinforcement learning (RL) is costly. Action-conditioned world models can provide imagined feedback by predicting visual consequences for candidate actions. However, a frozen world model may become less reliable as the policy evolves. In this paper, we introduce \method{}, a post-training framework that casts the coupled adaptation of a world model and an action model (policy) as task-specific recursive self-improvement (RSI). Each cycle alternates two updates. The world model evaluates policy actions through imagined outcomes, providing comparative feedback for group-relative policy optimization (GRPO). The improved policy then constructs a grounded self-curriculum, selecting expert-consistent action--video pairs by behavioral novelty and prediction error. The refined world model supplies feedback for the next policy update, closing the recursive self-improvement loop. Experiments show that \method{} outperforms training baselines and prior methods, validating the proposed recursive self-improvement loop between the policy and world model. Real-world trials further demonstrate its practical applicability.
\end{abstract}

\section{Introduction}

\begin{figure}[t]
  \centering
  \includegraphics[width=0.98\columnwidth]{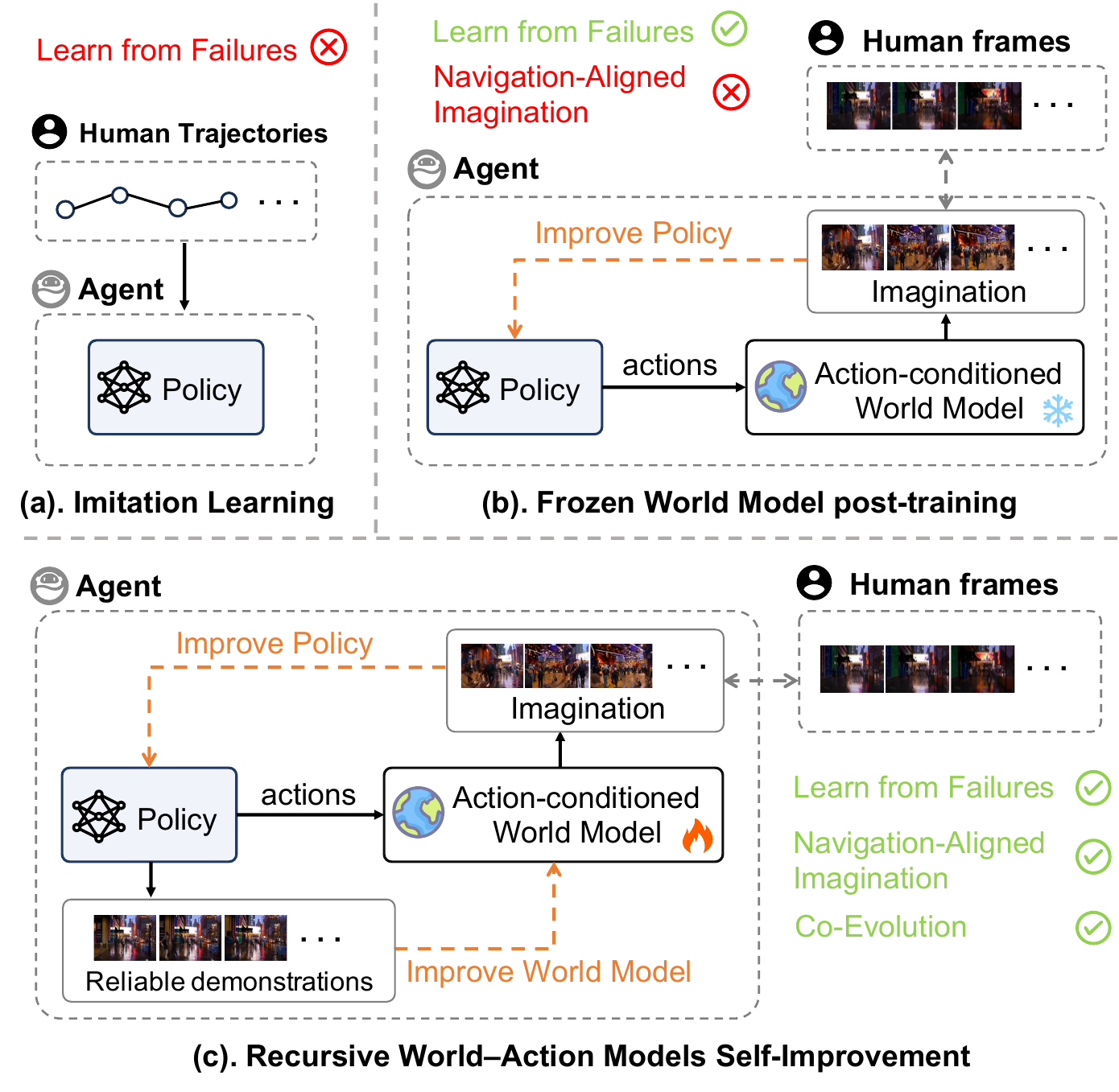}
  \caption{\textbf{From imitation learning to recursive world--action models self-improvement.} (a) Imitation learning trains the policy from expert trajectories but provides limited supervision for policy-induced failures. (b) A frozen world model enables learning from imagined outcomes, but its predictions become misaligned with the evolving policy. (c) \method{} closes the training loop: imagined outcomes improve the policy, while grounded policy-induced samples refine the world model for subsequent policy learning.}
  \label{fig:intro}
  \vspace{-18pt}
\end{figure}

Long-horizon urban navigation requires an agent to reach a language-specified destination from street-level observations and recent motion history. At each step, the agent must translate the global goal into a local decision, recover from mistakes, and determine when to stop. Existing methods commonly learn this behavior by imitating expert trajectories~\cite{liu2025citywalker,mei2026urbannav}. Although imitation learning (IL) provides strong initialization, it mainly supervises successful expert actions and cannot learn from policy-induced failures, as shown in Fig.~\ref{fig:intro}(a). This limitation is particularly consequential in urban navigation, where one inaccurate local motion changes the next observation and can compound over subsequent decisions~\cite{zhu2026wmpo}. Reinforcement learning (RL) can learn from such failures, but collecting long-horizon trial-and-error interaction is costly. Navigation post-training therefore needs outcome feedback without requiring every candidate action to be physically executed.

Action-conditioned world models serve as learned simulators for policy improvement. 
Navigation world models (NWM)~\cite{bar2025nwm} and recent world action models~\cite{zhu2026wmpo,ye2026dreamzero,yuan2026fastwam} predict future observations conditioned on candidate actions. Their imagined outcomes allow a policy to compare local decisions before execution. However, a world model used for policy optimization should not remain frozen, as illustrated in Fig.~\ref{fig:intro}(b). As the policy improves, the frozen world model does not adapt to the evolving distribution of urban navigation behaviors. Inaccurate predictions can then distort subsequent policy updates, making the frozen model a bottleneck of model-based post-training.

We therefore seek a training framework that adapts the world model alongside the policy. Recursive self-improvement (RSI) motivates a training process in which each improvement stage changes the conditions for the next~\cite{schmidhuber2007godel}. We bring this principle to navigation post-training by coupling policy optimization with world-model adaptation. The improved policy determines which demonstrations the world model should learn next, while the refined world model provides updated feedback for subsequent policy learning. Self-improvement therefore emerges from a recursive training loop between the policy and world model.

We introduce \method{} (\textbf{R}ecursive Self-\textbf{I}mprovement of \textbf{W}orld--\textbf{A}ction Models for \textbf{Nav}igation), a post-training framework that recursively improves a world model and an action model (policy). Each cycle alternates two coupled updates, as shown in Fig.~\ref{fig:intro}(c). First, the action-conditioned world model predicts visual outcomes for multiple policy actions. Their relative quality provides imagined feedback for Group Relative Policy Optimization (GRPO)~\cite{shao2024deepseekmath}. Second, the improved policy constructs candidate action--video pairs for adapting the world model to the policy's evolving behavior distribution. To maintain valid supervision from the offline dataset, we retain candidates that remain locally consistent with recorded expert motion. A grounded self-curriculum then prioritizes these samples according to behavioral novelty and current prediction error. The refined world model supplies the imagined feedback for the next policy update, closing the recursive training loop.

Our contributions are concluded as follows:
\begin{itemize}
  \item We introduce \method{}, a post-training framework that recursively improves a navigation policy and a visual world model.
  \item We develop a self-curriculum that filters policy proposals by expert consistency and prioritizes accepted action--video pairs by behavioral novelty and prediction error.
  \item Experiments demonstrate consistent improvements across seen and unseen environments, with real-world deployment further validating practical effectiveness.
\end{itemize}

\section{Related Work}

\subsection{Real-World Navigation}

Real-world navigation policies are learned from diverse robot trajectories and goal-conditioned supervision~\cite{shah2022recon,shah2023vint,sridhar2024nomad}. 
CityWalker~\cite{liu2025citywalker} scales urban navigation learning with large collections of walking and driving videos.
\dataset{}~\cite{mei2026urbannav} aligns city-scale human trajectories with landmark-grounded language.
OmniVLA~\cite{hirose2026omnivla} learns a shared navigation policy across image, language, and mixed goal modalities.
Other works explore generative guidance: NoMaD~\cite{sridhar2024nomad} uses a goal-masked diffusion policy, whereas VENTURA~\cite{zhang2025ventura} generates instruction-conditioned visual paths for downstream control.
These methods primarily improve navigation through data, supervision, and policy design. Our work instead focuses on post-training.

\subsection{World Models for Policy Learning}

World models support planning and policy learning through learned future dynamics~\cite{chua2018pets,janner2019mbpo}. PlaNet~\cite{hafner2019planet} and Dreamer~\cite{hafner2020dreamer} established latent imagination for policy learning, with later works scaling to broader domains~\cite{hafner2023dreamerv3,hansen2024tdmpc2} and high-dimensional visual observations~\cite{micheli2023iris,hansen2023modem}.
Recent world models extend to navigation and embodied control~\cite{zhou2025dinowm,yao2025navmorph}, while increasingly connect visual prediction with policy optimization~\cite{zhu2026wmpo,yang2026rise}. NWM~\cite{bar2025nwm} ranks navigation trajectories using controllable video prediction, whereas Fast-WAM~\cite{yuan2026fastwam} jointly models video and actions to improve policy learning without explicit future generation.
Rather than treating the world model as a fixed source of imagined feedback, we study how the world model and policy can continually adapt to one another as policy behavior evolves.

\subsection{Recursive Self-Improvement}

Recursive self-improvement traditionally studies systems that improve their own problem-solving process~\cite{schmidhuber2007godel}. POET~\cite{wang2019poet} co-evolves environment--agent pairs, generates new challenges, and transfers solutions across environments.
Model-based policy learning faces a related problem at the level of learned dynamics. Uncertainty-aware prediction and short rollouts reduce exploitation of model errors~\cite{chua2018pets,janner2019mbpo}, while conservative objectives penalize uncertain transitions~\cite{yu2020mopo,kidambi2020morel}. PDML~\cite{wang2023pdml} adapts model learning toward the current policy distribution. Other approaches prioritize experience through learning progress, novelty, or curiosity~\cite{kim2020activewm,sekar2020plan2explore,kauvar2023curiousreplay}.
In this paper, we connect these directions: the world model improves the policy through imagined feedback, and the evolving policy guides the next world-model update. This closed loop enables the two components to improve one another recursively.

\section{Method}
\label{sec:method}

\begin{figure*}[t]
  \centering
  \includegraphics[width=0.98\textwidth]{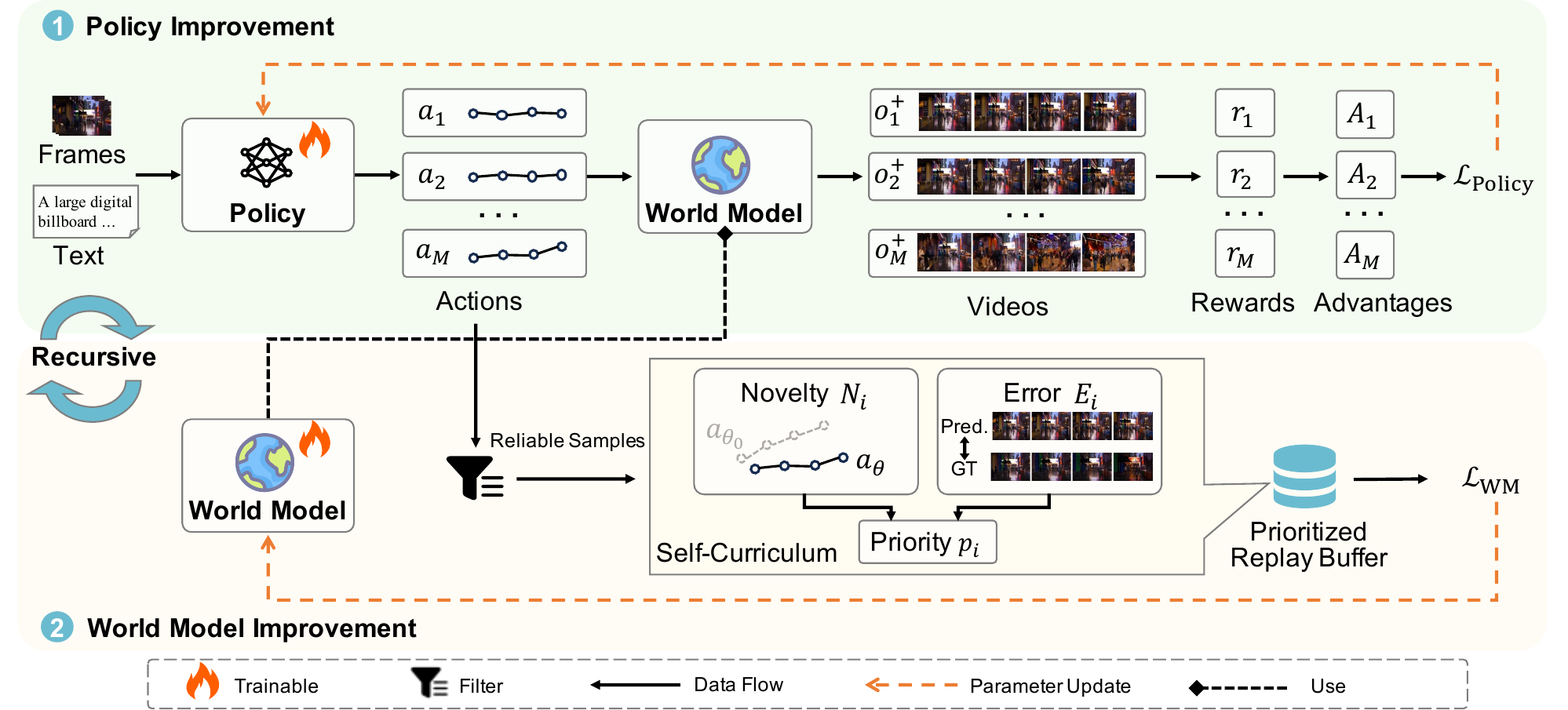}
  \caption{\textbf{Recursive self-improvement loop in \method.} (1) The current world model imagines action consequences that improve the policy. (2) The improved policy constructs a grounded self-curriculum that improves the next world model. Repeating these updates recursively changes both components and the feedback exchanged between them. The flame marks the component optimized in each phase.}
  \label{fig:overview}
  \vspace{-8pt}
\end{figure*}

We introduce \method{}, a post-training framework that recursively couples a navigation policy with an action-conditioned world model.
As shown in Fig.~\ref{fig:overview}, each cycle alternates between policy optimization and world-model adaptation. First, the current world model predicts the visual outcomes of candidate actions, providing a basis for comparing actions and improving the policy. Second, the improved policy generates candidate actions, and those consistent with expert motion are paired with recorded future frames to build the policy-induced replay pool. The grounded self-curriculum prioritizes these pairs by behavioral novelty and current prediction error for world-model adaptation. The refined world model then supplies imagined feedback for the next policy update, closing the recursive self-improvement loop.

\subsection{Overall}

We denote the offline \dataset{} dataset~\cite{mei2026urbannav} by $\mathcal D=\{(s_t,a_t^{\mathcal D},o_{t+1:t+T}^{\mathcal D})\}_t$, where $a_t^{\mathcal D}$ is the recorded expert action at time $t$ and $o_{t+1:t+T}^{\mathcal D}$ contains its subsequent $T$ RGB frames.
A state $s_t=(o_{t-C+1:t},q_{t-C+1:t},\ell)$ contains the most recent $C$ RGB observations $o$, their poses $q$, and a language goal $\ell$.
The policy $\pi_\theta(a\mid s)$ defines a distribution over an $H$-waypoint action $a=[p_1,\ldots,p_H]\in\R^{H\times2}$ in the current camera frame, where each $p_j\in\R^2$ is a planar waypoint; a separate head predicts the stop probability.

The action-conditioned \wm{} world model~\cite{nvidia2026cosmos3} predicts the visual consequence of a proposed action as
\begin{equation}
  \hat{o}_{t+1:t+T}=f_\phi(o_t,\mathcal A(a),\ell),
  \label{eq:wm}
\end{equation}
where $\mathcal A$ converts the waypoint chunk into the relative-pose sequence consumed by \wm{}.
The supervised policy $\pi_{\theta_0}$ and pretrained world model $f_{\phi_0}$ initialize the recursion. During recursion stage $k$, $f_{\phi_k}$ remains frozen while the policy improves by comparing imagined outcomes for actions proposed from the same state. The improved policy then supplies expert-consistent action--video pairs to the policy-induced replay pool. The grounded self-curriculum selects samples from this pool to update $f_{\phi_k}$ to $f_{\phi_{k+1}}$.
The recorded expert action $a_t^{\mathcal D}$ anchors future supervision, while $\pi_{\theta_0}$ provides a fixed reference for  behavioral novelty. Every policy-proposed action can inform policy optimization, but only expert-consistent actions are paired with recorded future frames and added to the policy-induced replay pool. The self-curriculum prioritizes these expert-consistent action--video pairs by behavioral novelty and current prediction error. The resulting $f_{\phi_{k+1}}$ supplies imagined feedback for the next policy-improvement phase. Repeating this recursion yields the final world model $f_{\phi_K}$.

\subsection{World Simulation for Policy Improvement}

\paragraph{Policy cold-start}
We first obtain a strong navigation policy through supervised fine-tuning (SFT), implemented as behavior cloning on expert trajectories from \dataset{}~\cite{mei2026urbannav}.
Frozen DINOv2 visual features~\cite{oquab2024dinov2}, CLIP text features~\cite{radford2021clip}, and pose embeddings are fused by a transformer waypoint head.
The corresponding SFT objective $\mathcal L_{\mathrm{SFT}}$ combines waypoint $L_1$ error, directional error, and stop-prediction binary cross-entropy on the expert data.
The resulting $\pi_{\theta_0}$ provides the initialization and a fixed behavior reference during post-training, but does not by itself compare alternative actions proposed by the learned policy.

\paragraph{Learning from imagination}
At state $s$, the current policy outputs a mean waypoint chunk $\mu_\theta(s)$.
We sample $M$ candidate actions $a_i\sim\mathcal N(\mu_\theta(s),\sigma^2I)$, where $\sigma$ controls the sampling scale, $I$ is the identity matrix, and $i\in\{1,\ldots,M\}$ indexes candidates.
The current world model, held fixed during this policy-update phase, predicts each candidate's visual outcome $\hat o_i^+$ through Eq.~\eqref{eq:wm}.
The navigation scoring function $R_{\mathrm{nav}}$ assigns each candidate a score $r_i=R_{\mathrm{nav}}(s,a_i,\hat{o}_i^+)$ that measures the quality of its imagined consequence.
Comparing candidates from the same state reduces sensitivity to absolute reward scale and directly identifies which local decision produces the better imagined outcome.
We perform online policy optimization in the world simulator using GRPO~\cite{shao2024deepseekmath}. States are drawn from the offline dataset, while candidate actions are sampled from the current policy and evaluated through imagined outcomes. We compute the group-relative advantage as
\begin{equation}
  A_i=\frac{r_i-\bar r}{\operatorname{std}(r_{1:M})+\epsilon_A},
  \label{eq:advantage}
\end{equation}
where $\bar r$ and $\operatorname{std}(r_{1:M})$ are the mean and standard deviation of the $M$ rewards, and $\epsilon_A>0$ ensures numerical stability.
Let $\pi_{\mathrm{old}}$ be the policy that sampled the candidate group and $\rho_i(\theta)=\pi_\theta(a_i\mid s)/\pi_{\mathrm{old}}(a_i\mid s)$ its likelihood ratio.
During post-training, we retain the waypoint and direction components of the supervised objective as an expert-anchor loss $\mathcal L_{\mathrm{exp}}$.
For the waypoint-action update, we optimize
\begin{align}
  \mathcal L_{\mathrm{Policy}}={}&-\frac{1}{M}\sum_{i=1}^{M}\min\!\left(
  \rho_i A_i,\operatorname{clip}(\rho_i,1-\eta,1+\eta)A_i\right) \notag\\
  &+\beta D_{\mathrm{KL}}\!\left(\pi_\theta(\cdot\mid s)\|\pi_{\theta_0}(\cdot\mid s)\right)
  +\lambda_{\mathrm{exp}}\mathcal L_{\mathrm{exp}},
  \label{eq:policy}
\end{align}
where $\eta$ is the ratio-clipping range, while $\beta$ and $\lambda_{\mathrm{exp}}$ weight the reference-policy Kullback-Leibler (KL) penalty and expert anchor, respectively.
The expert anchor compares the current prediction with recorded expert waypoints, whereas the KL term constrains the action distribution relative to the frozen initial policy $\pi_{\theta_0}$.
Thus, the group-relative term learns from imagined action consequences, while clipping, the KL penalty, and the expert anchor prevent unstable drift from the supervised policy.

\subsection{Curriculum Learning for World Model Improvement}

\paragraph{Policy-induced Replay}
The improved policy determines the behaviors on which the next world-model update should focus.
However, not every policy-proposed action forms a reliable correspondence with the recorded future $o_{t+1:t+T}^{\mathcal D}$. Actions that deviate substantially from $a_t^{\mathcal D}$ lack valid supervision.
We therefore retain only policy-proposed actions that remain locally consistent with recorded expert motion.
Following \dataset{}~\cite{mei2026urbannav}, we use average distance error (ADE), maximum average distance error (MADE), average orientation error (AOE), and maximum average orientation error (MAOE) as action-to-expert discrepancies.
Using $\mathcal M=\{\mathrm{ADE},\mathrm{MADE},\mathrm{AOE},\mathrm{MAOE}\}$ to index these metrics, we define the expert-consistency score
\begin{equation}
  Q_i=\frac{\sum_{m\in\mathcal M}w_m
  \exp\!\left[-d_m(a_i,a_t^{\mathcal D})/\tau_m\right]}
  {\sum_{m\in\mathcal M}w_m},
  \label{eq:reliability}
\end{equation}
where $d_m(a_i,a_t^{\mathcal D})$ is the discrepancy from the recorded expert action under metric $m$, $w_m$ is its nonnegative weight, and $\tau_m>0$ is a metric-specific scale.
The score $Q_i\in[0,1]$ summarizes the candidate's consistency with recorded expert motion across the distance and orientation measures.
We admit a candidate only when its rollout is valid, $Q_i\geq q_{\min}$, and the individual maximum-error tolerances are satisfied, where $q_{\min}$ is the expert-consistency threshold.
Each retained action and its recorded future form an expert-consistent action--video pair, which serves as a reliable sample for world-model training.
Policy-proposed actions that fail the check are skipped for world-model training but remain useful for policy optimization.
The filter prevents mismatched action--video pairs while allowing the policy phase to explore more broadly.
Each admitted pair $(s,a_i,o^{+\mathcal D})$ enters the policy-induced replay pool $\mathcal R_k$.

\paragraph{Grounded self-curriculum}
Uniform replay treats all accepted samples equally, regardless of their behavioral novelty or prediction error. The improved policy instead constructs a grounded self-curriculum from expert-consistent action--video pairs. The self-curriculum prioritizes each pair by behavioral novelty relative to $\pi_{\theta_0}$ and current prediction error under $f_{\phi_k}$.
For candidate $a_i$, we measure its behavioral deviation from the initial policy mean as $D_i=\lVert a_i-\mu_{\theta_0}(s)\rVert_2/\sqrt{2H}$.
Let $e_i$ be the mean pixel $L_1$ error between temporally aligned frames from the predicted and recorded futures.
We map these quantities to bounded novelty and prediction-error scores:
\begin{equation}
  N_i=1-\exp(-D_i/\tau_N), \qquad
  E_i=1-\exp(-\tau_E e_i),
  \label{eq:curriculum_scores}
\end{equation}
where $\tau_N>0$ and $\tau_E>0$ control the respective scales.
The replay priority is
\begin{equation}
  p_i=Q_i\left(\epsilon_p+\lambda_NN_i+\lambda_EE_i\right),
  \label{eq:priority}
\end{equation}
where $\lambda_N$ and $\lambda_E$ balance novelty and error, while $\epsilon_p>0$ preserves nonzero priority.
At each world-model update, newly admitted and historical pairs compete within the same replay pool. We sample without replacement in proportion to $p_i^\alpha$, where $\alpha\geq0$ controls the strength of prioritization. The sampled batch defines the world-model objective
\begin{equation}
  \mathcal L_{\mathrm{WM}}(\phi)=
  \mathbb E_{(s,a_i,o^{+\mathcal D})\sim\mathcal B_p}
  \!\left[\mathcal L_{\mathrm{video}}
  \big(f_\phi(o_t,\mathcal A(a_i),\ell),o^{+\mathcal D}\big)\right],
  \label{eq:wmloss}
\end{equation}
where $\mathcal B_p\subset\mathcal R_k$ denotes the sampled replay batch and $o^{+\mathcal D}$ is the recorded future paired with $a_i$. The term $\mathcal L_{\mathrm{video}}$ denotes the native conditional video-model objective of \wm{}~\cite{nvidia2026cosmos3}.
The three factors serve distinct roles. The score $Q_i$ weights each pair by expert consistency, while $N_i$ and $E_i$ prioritize behavioral novelty and current prediction error. Together, they focus adaptation on expert-consistent pairs that differ from the initial policy's behavior or remain poorly predicted by the current world model.
After adaptation, $f_{\phi_{k+1}}$ evaluates subsequent policy-proposed actions and supplies updated imagined feedback for policy improvement. This reciprocal exchange closes the recursive self-improvement loop.

\begin{table*}[t]
\centering
\caption{Navigation performance on the UrbanNav test-seen and test-unseen splits. Values are reported as mean $\pm$ standard deviation.}
\label{tab:main}
\scriptsize
\setlength{\tabcolsep}{5.0pt}
\begin{tabular}{@{}lcccccc@{}}
\toprule
\multicolumn{7}{c}{Test Seen}\\
\cmidrule(lr){1-7}
Training framework & SR (\%)$\uparrow$ & SPL (\%)$\uparrow$ & NE$\downarrow$ & nDTW (\%)$\uparrow$ & SDTW (\%)$\uparrow$ & OSR (\%)$\uparrow$\\
\midrule
Imitation learning
& 70.91 & 70.55 & 3.8427 & 83.81 & 69.16 & 71.18 \\
PPO~\cite{schulman2017ppo}
& \meanstd{72.04}{0.06} & \meanstd{71.40}{0.09} & \meanstd{3.8634}{0.0136} & \meanstd{84.02}{0.08} & \meanstd{70.29}{0.06} & \meanstd{72.36}{0.07}
\\
GRPO~\cite{shao2024deepseekmath}
& \meanstd{75.11}{0.75} & \meanstd{74.45}{0.75} & \meanstd{3.7012}{0.1803} & \meanstd{86.40}{0.47} & \meanstd{73.37}{0.76} & \meanstd{75.32}{0.81}
\\
PPO-WM
& \meanstd{78.90}{0.18} & \meanstd{78.25}{0.19} & \meanstd{2.9530}{0.0647} & \meanstd{88.06}{0.13} & \meanstd{77.36}{0.18} & \meanstd{79.39}{0.20}
\\
GRPO-WM
& \meanstd{83.35}{1.25} & \meanstd{81.27}{1.38} & \meanstd{2.7059}{0.1858} & \meanstd{90.55}{0.73} & \meanstd{81.66}{1.29} & \meanstd{84.31}{1.17}
\\
\midrule
\method{} (ours)
& \bestmeanstd{89.92}{0.52} & \bestmeanstd{86.87}{0.77} & \bestmeanstd{1.7842}{0.1408} & \bestmeanstd{94.21}{0.26} & \bestmeanstd{88.28}{0.56} & \bestmeanstd{91.88}{0.30}
\\
\midrule
\multicolumn{7}{c}{Test Unseen}\\
\cmidrule(lr){1-7}
Training framework & SR (\%)$\uparrow$ & SPL (\%)$\uparrow$ & NE$\downarrow$ & nDTW (\%)$\uparrow$ & SDTW (\%)$\uparrow$ & OSR (\%)$\uparrow$\\
\midrule
Imitation learning
& 67.88 & 66.03 & 4.4537 & 80.85 & 64.99 & 68.16 \\
PPO~\cite{schulman2017ppo}
& \meanstd{67.94}{0.09} & \meanstd{65.65}{0.16} & \meanstd{4.3768}{0.0434} & \meanstd{80.76}{0.05} & \meanstd{65.04}{0.08} & \meanstd{68.38}{0.07} \\
GRPO~\cite{shao2024deepseekmath}
& \meanstd{70.63}{1.10} & \meanstd{68.49}{1.06} & \meanstd{4.5795}{0.0894} & \meanstd{82.34}{0.54} & \meanstd{67.82}{1.13} & \meanstd{70.91}{1.07} \\
PPO-WM
& \meanstd{75.05}{0.21} & \meanstd{72.44}{0.19} & \meanstd{3.5912}{0.0462} & \meanstd{84.96}{0.15} & \meanstd{72.27}{0.22} & \meanstd{75.56}{0.21} \\
GRPO-WM
& \meanstd{80.31}{0.94} & \meanstd{76.09}{1.05} & \meanstd{3.3314}{0.1014} & \meanstd{87.77}{0.45} & \meanstd{77.62}{0.92} & \meanstd{81.01}{0.82} \\
\midrule
\method{} (ours)
& \bestmeanstd{88.25}{0.43} & \bestmeanstd{82.59}{0.25} & \bestmeanstd{2.0429}{0.0815} & \bestmeanstd{91.95}{0.16} & \bestmeanstd{85.51}{0.39} & \bestmeanstd{89.37}{0.44} \\
\bottomrule
\end{tabular}
\end{table*}

\section{Experiments}

We test our \method{} through three questions. First, can a learned visual world model improve the navigation policy beyond policy-only training? Second, does policy-guided world-model refinement improve over a frozen world model? Third, does the world model become more accurate on behaviors from the improved policy? Controlled comparisons, ablations, and world-model analyses address these questions in sequence.
A physical-robot deployment evaluates transfer beyond offline benchmarking.

\subsection{Experimental Setup}

\paragraph{Dataset and evaluation}
We evaluate on the official train, test-seen, and test-unseen splits of \dataset{}~\cite{mei2026urbannav}.
All training frameworks receive the same recent-frame, pose-history, and language goals and predict eight local waypoints plus a stop probability.
Every framework uses the same data split, episode list, trajectory mapping, and 3-m success radius. Unless otherwise stated, results are means and standard deviations over three random seeds.
Success Rate (SR) measures final goal attainment, Oracle Success Rate (OSR) records whether any visited pose reaches the goal region, and Navigation Error (NE) measures final goal distance. Success weighted by Path Length (SPL) evaluates path efficiency~\cite{anderson2018navigationeval}; normalized Dynamic Time Warping (nDTW) and success-weighted nDTW (SDTW) evaluate trajectory fidelity~\cite{ilharco2019dtw}. Higher values are better except for NE.

\paragraph{Baselines}
The controlled comparison fixes the policy architecture~\cite{mei2026urbannav}, initialization, data, interface, and evaluation protocol while changing the post-training framework.
Imitation learning~\cite{mei2026urbannav} is the common policy trained on expert trajectories.
PPO~\cite{schulman2017ppo} and GRPO~\cite{shao2024deepseekmath} are policy-only baselines that optimize the same navigation objective without action-conditioned visual feedback.
PPO-WM and GRPO-WM use the corresponding policy optimizer with imagined consequences from a frozen \wm{}~\cite{nvidia2026cosmos3}.
\method{} uses GRPO-based policy improvement and additionally adapts \wm{} with the proposed grounded self-curriculum.

\paragraph{Implementation details}
The policy predicts eight waypoints over eight seconds and compares four candidates at each state; \wm{} predicts 17 frames at 2~fps from 16 relative-pose actions.
Policy training uses a learning rate of $10^{-5}$ and clipping range $\eta=0.20$.
The self-curriculum uses expert-consistency threshold $q_{\min}=0.35$ and equal novelty and prediction-error weights $\lambda_N=\lambda_E=0.50$.
The world model is updated every 500 policy steps using at most 500 replay examples.
Training runs on four NVIDIA H200 GPUs. Real-world deployment uses a custom-built mobile robotic platform equipped with a forward-facing camera that provides RGB observations.

\begin{table*}[t]
\centering
\caption{Comparison with representative prior approaches on the UrbanNav test-seen and test-unseen splits. Values are reported as mean $\pm$ standard deviation.}
\label{tab:navigation_methods}
\scriptsize
\setlength{\tabcolsep}{3.5pt}
\begin{tabular}{@{}clcccccc@{}}
\toprule
\multicolumn{8}{c}{Test Seen}\\
\cmidrule(lr){1-8}
Category & Method & SR (\%)$\uparrow$ & SPL (\%)$\uparrow$ & NE$\downarrow$ & nDTW (\%)$\uparrow$ & SDTW (\%)$\uparrow$ & OSR (\%)$\uparrow$\\
\cmidrule(lr){1-8}
\multirow{3}{*}{\shortstack{Imitation\\Learning}} & UrbanNav~\cite{mei2026urbannav} & \meanstd{60.36}{5.39} & \meanstd{59.75}{5.34} & \meanstd{5.3958}{0.6689} & \meanstd{76.86}{3.50} & \meanstd{58.22}{5.34} & \meanstd{60.48}{5.36} \\
& NoMaD~\cite{sridhar2024nomad} & \secondmeanstd{82.35}{0.78} & \secondmeanstd{77.59}{0.55} & \secondmeanstd{2.5532}{0.0608} & \meanstd{90.15}{0.19} & \secondmeanstd{77.97}{0.79} & \secondmeanstd{82.47}{0.76} \\
& OmniVLA~\cite{hirose2026omnivla} & \meanstd{77.49}{1.66} & \meanstd{77.25}{1.66} & \meanstd{3.1856}{0.1415} & \secondmeanstd{90.25}{0.47} & \meanstd{73.63}{1.72} & \meanstd{77.66}{1.62} \\
\cmidrule(lr){1-8}
\multirow{2}{*}{WM-based} & Fast-WAM~\cite{yuan2026fastwam} & \meanstd{58.48}{0.75} & \meanstd{39.57}{0.18} & \meanstd{8.4899}{0.0723} & \meanstd{64.77}{0.20} & \meanstd{50.62}{0.56} & \meanstd{62.07}{0.31} \\
& NWM~\cite{bar2025nwm} & \meanstd{71.26}{0.21} & \meanstd{70.67}{0.21} & \meanstd{3.7117}{0.0200} & \meanstd{83.88}{0.05} & \meanstd{69.45}{0.20} & \meanstd{71.63}{0.20} \\
\cmidrule(lr){1-8}
\shortstack{Recursive\\Self-Improvement} & \tabvcenter{\method{} (ours)} & \tabvcenter{\bestmeanstd{89.92}{0.52}} & \tabvcenter{\bestmeanstd{86.87}{0.77}} & \tabvcenter{\bestmeanstd{1.7842}{0.1408}} & \tabvcenter{\bestmeanstd{94.21}{0.26}} & \tabvcenter{\bestmeanstd{88.28}{0.56}} & \tabvcenter{\bestmeanstd{91.88}{0.30}} \\
\midrule
\multicolumn{8}{c}{Test Unseen}\\
\cmidrule(lr){1-8}
Category & Method & SR (\%)$\uparrow$ & SPL (\%)$\uparrow$ & NE$\downarrow$ & nDTW (\%)$\uparrow$ & SDTW (\%)$\uparrow$ & OSR (\%)$\uparrow$\\
\cmidrule(lr){1-8}
\multirow{3}{*}{\shortstack{Imitation\\Learning}} & UrbanNav~\cite{mei2026urbannav} & \meanstd{60.99}{4.01} & \meanstd{59.45}{3.99} & \meanstd{5.3680}{0.3845} & \meanstd{76.03}{2.91} & \meanstd{57.93}{3.99} & \meanstd{61.18}{4.02} \\
& NoMaD~\cite{sridhar2024nomad} & \secondmeanstd{84.80}{0.52} & \meanstd{79.85}{0.30} & \secondmeanstd{2.3520}{0.0656} & \meanstd{90.77}{0.21} & \secondmeanstd{80.37}{0.55} & \secondmeanstd{85.09}{0.51} \\
& OmniVLA~\cite{hirose2026omnivla} & \meanstd{81.11}{1.31} & \secondmeanstd{80.80}{1.29} & \meanstd{2.9283}{0.0947} & \secondmeanstd{90.82}{0.42} & \meanstd{77.05}{1.44} & \meanstd{81.17}{1.33} \\
\cmidrule(lr){1-8}
\multirow{2}{*}{WM-based} & Fast-WAM~\cite{yuan2026fastwam} & \meanstd{55.25}{0.01} & \meanstd{35.00}{0.33} & \meanstd{9.2896}{0.0221} & \meanstd{61.56}{0.08} & \meanstd{47.16}{0.03} & \meanstd{59.75}{0.50} \\
& NWM~\cite{bar2025nwm} & \meanstd{67.99}{0.08} & \meanstd{65.93}{0.07} & \meanstd{4.4130}{0.0604} & \meanstd{80.68}{0.04} & \meanstd{65.06}{0.08} & \meanstd{68.46}{0.09} \\
\cmidrule(lr){1-8}
\shortstack{Recursive\\Self-Improvement} & \tabvcenter{\method{} (ours)} & \tabvcenter{\bestmeanstd{88.25}{0.43}} & \tabvcenter{\bestmeanstd{82.59}{0.25}} & \tabvcenter{\bestmeanstd{2.0429}{0.0815}} & \tabvcenter{\bestmeanstd{91.95}{0.16}} & \tabvcenter{\bestmeanstd{85.51}{0.39}} & \tabvcenter{\bestmeanstd{89.37}{0.44}} \\
\bottomrule
\end{tabular}
\vspace{-10pt}
\end{table*}

\subsection{Comparative Experiments}

\subsubsection{Comparison with Training Frameworks}

Table~\ref{tab:main} compares all training frameworks under the same policy architecture, initialization, data, and evaluation protocol.
This controlled study tests whether world-model adaptation adds value beyond the choice of policy optimizer and the availability of imagined feedback.

Policy-only post-training provides limited gains from the shared imitation-learned initialization. On test-unseen, PPO changes SR/SPL by $+0.06\%/-0.38\%$, whereas GRPO improves them by $2.75\%/2.46\%$.
Frozen-world-model feedback produces a larger change: PPO-WM improves SR/SPL over PPO by $7.11\%/6.79\%$, and GRPO-WM improves them over GRPO by $9.68\%/7.60\%$.
These controlled comparisons isolate imagined visual consequences as a useful training signal even before world-model adaptation.

Closing the recursive feedback loop provides a further gain over frozen world-model feedback. Relative to GRPO-WM, \method{} improves SR/SPL by $6.57\%/5.60\%$ on test-seen and $7.94\%/6.50\%$ on test-unseen, achieving the best performance across all reported metrics on both splits.
On test-unseen, it also reduces NE from 3.3314 to 2.0429 and raises nDTW, SDTW, and OSR from 87.77\%/77.62\%/81.01\% to 91.95\%/85.51\%/89.37\%.
The agreement across success, path efficiency, route fidelity, and oracle success indicates that policy-guided world-model adaptation improves the complete rollout rather than only its stopping decision.

\subsubsection{Comparison with Prior Approaches}

Table~\ref{tab:navigation_methods} compares \method{} with imitation-learning and world-model-based methods on the \dataset{}~\cite{mei2026urbannav} test-seen and test-unseen splits.
All systems are evaluated under the same closed-loop protocol.
We use official checkpoints when available and otherwise adapt the goal or action interface. NoMaD receives text conditioning and a stop head, OmniVLA is fine-tuned through LoRA and its continuous action head, and NWM~\cite{bar2025nwm} reranks four policy candidates by visual-language goal alignment. UrbanNav~\cite{mei2026urbannav} and Fast-WAM~\cite{yuan2026fastwam} retain their trained checkpoints and native action interfaces.

Among the baselines, NWM improves test-unseen SR/SPL over UrbanNav by $7.00\%/6.48\%$ while reducing NE by $0.9550$, confirming the value of predictive visual feedback in the common pipeline. \method{} achieves the best performance across all six metrics on both test splits. Relative to NoMaD, the strongest prior approach in SR, it improves test-unseen SR/SPL by $3.45\%/2.74\%$, reduces NE by $0.3091$, and raises SDTW/OSR by $5.14\%/4.28\%$. Relative to OmniVLA, which has the strongest prior SPL, \method{} improves SR/SPL by $7.14\%/1.79\%$ and nDTW by $1.13\%$.
Among the predictive-model baselines, \method{} exceeds NWM in SR/SPL by $20.26\%/16.66\%$ and reduces NE from $4.4130$ to $2.0429$. The advantage over both learned policies and predictive-model reranking supports recursive self-improvement as a training effect rather than a consequence of a single baseline design.

\subsection{Ablation Study}

We evaluate world-model updates, policy-induced replay, and the grounded self-curriculum, as shown in Table~\ref{tab:ablation}. World-model updates use recorded expert action--video pairs by default (static expert replay). Policy-induced replay instead uses expert-consistent policy actions paired with recorded future frames. The grounded self-curriculum prioritizes these pairs by behavioral novelty and prediction error; otherwise, they are sampled uniformly.

\begin{table}[t]
\centering
\caption{Ablation study on the UrbanNav test-unseen split. Values are reported as mean $\pm$ standard deviation.}
\label{tab:ablation}
\footnotesize
\setlength{\tabcolsep}{2.5pt}
\begin{tabular}{@{}ccccc@{}}
\toprule
WM & Policy-induced & Grounded & \multirow{2}{*}{SR (\%)$\uparrow$} & \multirow{2}{*}{SPL (\%)$\uparrow$}\\
update & replay & self-curriculum & & \\
\midrule
$\times$ & $\times$ & $\times$ & \meanstd{80.31}{0.94} & \meanstd{76.09}{1.05}\\
$\checkmark$ & $\times$ & $\times$ & \meanstd{78.23}{1.91} & \meanstd{74.72}{1.92}\\
$\checkmark$ & $\checkmark$ & $\times$ & \meanstd{84.83}{0.33} & \meanstd{80.11}{0.32}\\
$\checkmark$ & $\checkmark$ & $\checkmark$ & \bestmeanstd{88.25}{0.43} & \bestmeanstd{82.59}{0.25}\\
\bottomrule
\end{tabular}
\vspace{-10pt}
\end{table}

With static expert replay, SR and SPL are 2.08\% and 1.37\% lower than those of the frozen-world-model baseline.
Compared with static expert replay, policy-induced replay with uniform sampling improves SR and SPL by 6.60\% and 5.39\%, respectively. This shows the benefit of using actions proposed by the improved policy for world-model training.
Adding the grounded self-curriculum to policy-induced replay further improves SR and SPL by 3.42\% and 2.48\%. This gain supports prioritizing expert-consistent samples by behavioral novelty and prediction error.
Combining policy-induced replay with the grounded self-curriculum improves SR and SPL by 7.94\% and 6.50\% over the frozen-world-model baseline. These results show that updating the world model with actions from the improved policy yields better navigation performance than freezing it or training it only on expert data.

\subsection{Evidence for Recursive Self-Improvement}

\begin{figure}[t]
    \centering
    \includegraphics[width=\columnwidth]{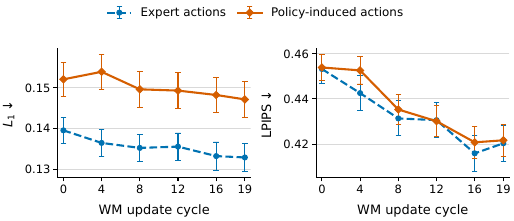}
    \caption{World-model prediction quality across recursive rounds. The left and right panels show $L_1$ and LPIPS on two fixed evaluation sets: expert actions and actions generated by the improved policy. Points are means over 128 samples and error bars denote standard error. Round 0 uses the initial world model, whereas the final round $K=19$ uses the final \method{} world model. Lower is better.}
    \label{fig:wm-evolution}
    \vspace{-14pt}
\end{figure}

\paragraph{World-model improvement across recursive rounds}
This experiment tests whether policy-guided world-model adaptation improves predictions for actions.
We compare the initial model, static expert adaptation, and the final \method{} world model on two fixed evaluation sets. One contains expert actions; the other contains actions generated by the improved policy near the end of training (policy-induced actions). Each set contains 128 state-action pairs excluded from all world-model updates.
Across the same eight-second horizon, we report frame-wise $L_1$ error and Learned Perceptual Image Patch Similarity (LPIPS); lower values indicate better pixel and perceptual fidelity. Table~\ref{tab:wm} reports endpoint quality, while Fig.~\ref{fig:wm-evolution} traces the same metrics across recursive rounds.

\begin{table}[t]
\centering
\caption{Endpoint world-model prediction quality on expert actions and actions generated by the improved policy.}
\label{tab:wm}
\footnotesize
\setlength{\tabcolsep}{3.5pt}
\begin{tabular}{@{}llcc@{}}
\toprule
\multirow{2}{*}{World model} & Evaluation & \multirow{2}{*}{$L_1\downarrow$} & \multirow{2}{*}{LPIPS$\downarrow$}\\
& actions & & \\
\midrule
\multirow{2}{*}{Initial $\phi_0$}
  & Expert & \meanstd{0.1396}{0.0032} & \meanstd{0.4532}{0.0065}\\
  & Policy-induced & \meanstd{0.1521}{0.0042} & \meanstd{0.4539}{0.0059}\\
\addlinespace[1pt]
\multirow{2}{*}{Static expert}
  & Expert & \meanstd{0.1352}{0.0034} & \meanstd{0.4265}{0.0079}\\
  & Policy-induced & \meanstd{0.1505}{0.0044} & \meanstd{0.4314}{0.0067}\\
\addlinespace[1pt]
\multirow{2}{*}{\method{} $\phi_K$}
  & Expert & \bestmeanstd{0.1329}{0.0034} & \bestmeanstd{0.4203}{0.0079}\\
  & Policy-induced & \bestmeanstd{0.1472}{0.0043} & \bestmeanstd{0.4217}{0.0070}\\
\bottomrule
\end{tabular}
\vspace{-4pt}
\end{table}

Table~\ref{tab:wm} and Fig.~\ref{fig:wm-evolution} jointly characterize world-model improvement across recursive rounds.
The final \method{} world model performs best on both distributions: relative to the initial model, it reduces L1/LPIPS by 4.8\%/7.3\% on expert actions and 3.2\%/7.1\% on actions generated by the improved policy.
Static expert adaptation improves both distributions. Policy-guided world-model adaptation further reduces L1/LPIPS by 1.7\%/1.5\% on expert actions and 2.2\%/2.2\% on policy-induced actions.
The figure traces the same evaluation metrics through the intermediate recursive rounds.
Despite small fluctuations between individual updates, both errors decrease overall on both action distributions, with most of the improvement retained through the final round.
Together, the endpoint comparison and the evolution curves show that policy-guided adaptation improves feedback for the evolving policy without sacrificing prediction quality on the original expert-action distribution.

\paragraph{Qualitative results}

\begin{figure*}[t!]
    \centering
    \includegraphics[width=0.86\textwidth]{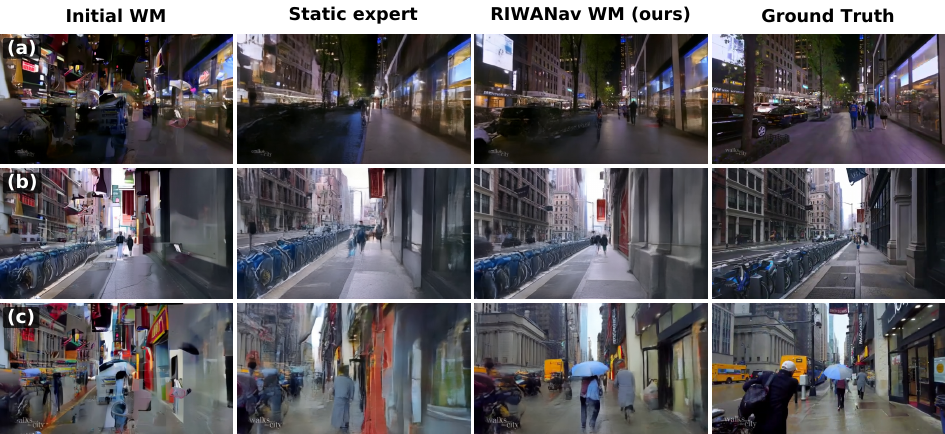}
    \caption{Qualitative evidence for recursive world-model improvement on actions generated by the improved policy. All three models receive the same observation, language goal, and action; the aligned recorded future is shown as reference. The rows depict night traffic, a bicycle-lined sidewalk, and a rainy sidewalk. In these cases, the \method{} world model after recursive updates better preserves traversable corridors and scene layout, with fewer long-horizon artifacts.}
    \label{fig:qualitative}
    \vspace{-12pt}
\end{figure*}

As shown in Fig.~\ref{fig:qualitative}, we provide matched qualitative comparisons on actions generated by the improved policy.
The initial world model frequently loses scene structure over the eight-second rollout, while static expert adaptation does not consistently resolve distortions under the improved policy distribution.
The final \method{} world model produces more coherent sidewalk boundaries, facade geometry, and arrangements of vehicles and pedestrians.
These examples are consistent with the L1 and LPIPS improvements on the evaluation sets in Table~\ref{tab:wm}: adaptation to policy-induced behaviors preserves scene structure more reliably for actions generated by the improved policy.

\subsection{Real-World Deployment}
\label{sec:real-world-deployment}

\begin{figure}[t]
    \centering
    \includegraphics[width=\columnwidth]{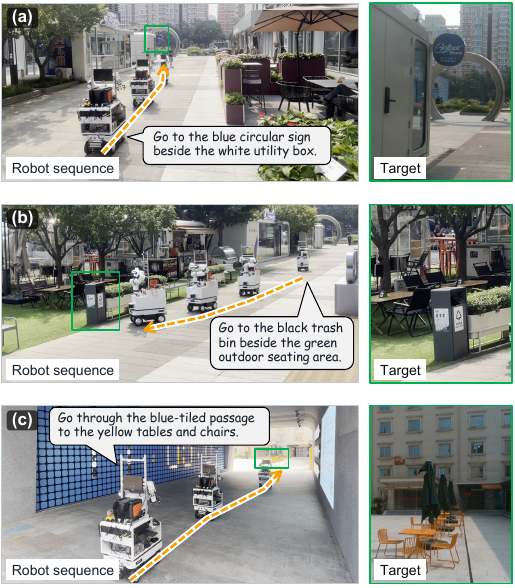}
    \caption{Real-world urban navigation with \method{}. Orange dashed arrows indicate schematic paths, and green boxes highlight the targets.}
    \label{fig:real-world-deployment}
    \vspace{-14pt}
\end{figure}

To assess real-world applicability, we deploy the policy trained with \method{} on a mobile robot in urban navigation scenarios. Given a natural-language instruction and onboard visual observations, the policy predicts waypoints to guide the robot toward the specified goal. Figure~\ref{fig:real-world-deployment} shows the resulting navigation sequences across different route layouts and lighting conditions. These qualitative trials demonstrate that the learned policy can guide a physical robot toward language-specified landmarks, supporting the practical feasibility of \method{} beyond offline evaluation.

\subsection{Discussion and Limitations}

The controlled comparisons and ablations clarify where \method{} gains arise. Imagined world-model feedback improves navigation beyond policy-only training, and policy-guided world-model adaptation provides further gains over a frozen world model. While these results support the recursive design, its current implementation remains task-specific and relies on a fixed offline dataset. Recorded futures can supervise only policy actions close to expert motion, limiting counterfactual exploration without new interaction data. Adapting a large video world model also increases memory and computation. Although real-world trials demonstrate practical feasibility, broader evaluations across cities, environments, and robot embodiments are needed.

\section{Conclusion}

We presented \method{}, a post-training framework for task-specific recursive self-improvement of a world model and a navigation policy in urban navigation. The current world model supplies imagined feedback to improve the policy, and the improved policy constructs a grounded self-curriculum for the next world-model update. Controlled experiments show that imagined feedback improves the policy and that policy-guided world-model adaptation provides further gains over a frozen world model. The adapted world model also improves prediction fidelity on actions generated by the improved policy. These results support recursive self-improvement of world--action models as a practical approach to urban navigation.


\balance
\bibliographystyle{IEEEtran}
\bibliography{references}

@inproceedings{mei2026urbannav,
  author    = {Yanghong Mei and Yirong Yang and Longteng Guo and Qunbo Wang and Ming-Ming Yu and Xingjian He and Wenjun Wu and Jing Liu},
  title     = {{UrbanNav}: Learning Language-Guided Embodied Urban Navigation from Web-Scale Human Trajectories},
  booktitle = {Proceedings of the AAAI Conference on Artificial Intelligence},
  volume    = {40},
  number    = {22},
  pages     = {18505--18513},
  year      = {2026},
  doi       = {10.1609/aaai.v40i22.38916}
}

@inproceedings{liu2025citywalker,
  author    = {Xinhao Liu and Jintong Li and Yicheng Jiang and Niranjan Sujay and Zhicheng Yang and Juexiao Zhang and John Abanes and Jing Zhang and Chen Feng},
  title     = {{CityWalker}: Learning Embodied Urban Navigation from Web-Scale Videos},
  booktitle = {Proceedings of the IEEE/CVF Conference on Computer Vision and Pattern Recognition},
  pages     = {6875--6885},
  year      = {2025}
}

@inproceedings{shah2022recon,
  author    = {Dhruv Shah and Benjamin Eysenbach and Nicholas Rhinehart and Sergey Levine},
  title     = {Rapid Exploration for Open-World Navigation with Latent Goal Models},
  booktitle = {Proceedings of the 5th Conference on Robot Learning},
  series    = {Proceedings of Machine Learning Research},
  volume    = {164},
  pages     = {674--684},
  year      = {2022}
}

@inproceedings{shah2023vint,
  author    = {Dhruv Shah and Ajay Sridhar and Nitish Dashora and Kyle Stachowicz and Kevin Black and Noriaki Hirose and Sergey Levine},
  title     = {{ViNT}: A Foundation Model for Visual Navigation},
  booktitle = {Proceedings of the 7th Conference on Robot Learning},
  series    = {Proceedings of Machine Learning Research},
  volume    = {229},
  pages     = {711--733},
  year      = {2023}
}

@inproceedings{bar2025nwm,
  author    = {Amir Bar and Gaoyue Zhou and Danny Tran and Trevor Darrell and Yann LeCun},
  title     = {Navigation World Models},
  booktitle = {Proceedings of the IEEE/CVF Conference on Computer Vision and Pattern Recognition},
  pages     = {15791--15801},
  year      = {2025}
}

@inproceedings{yao2025navmorph,
  author    = {Xuan Yao and Junyu Gao and Changsheng Xu},
  title     = {{NavMorph}: A Self-Evolving World Model for Vision-and-Language Navigation in Continuous Environments},
  booktitle = {Proceedings of the IEEE/CVF International Conference on Computer Vision},
  pages     = {5536--5546},
  year      = {2025}
}

@inproceedings{zhu2026wmpo,
  title     = {{WMPO}: World Model-based Policy Optimization for Vision-Language-Action Models},
  author    = {Zhu, Fangqi and Yan, Zhengyang and Hong, Zicong and Shou, Quanxin and Ma, Xiao and Guo, Song},
  booktitle = {International Conference on Learning Representations},
  volume    = {2026},
  pages     = {62486--62502},
  year      = {2026}
}

@article{yuan2026fastwam,
  author  = {Tianyuan Yuan and Zibin Dong and Yicheng Liu and Hang Zhao},
  title   = {{Fast-WAM}: Do World Action Models Need Test-time Future Imagination?},
  journal = {arXiv preprint arXiv:2603.16666},
  year    = {2026}
}

@inproceedings{sridhar2024nomad,
  author    = {Ajay Sridhar and Dhruv Shah and Catherine Glossop and Sergey Levine},
  title     = {{NoMaD}: Goal Masked Diffusion Policies for Navigation and Exploration},
  booktitle = {2024 IEEE International Conference on Robotics and Automation (ICRA)},
  pages     = {63--70},
  doi       = {10.1109/ICRA57147.2024.10610665},
  year      = {2024},
  url       = {https://arxiv.org/abs/2310.07896}
}

@inproceedings{hirose2026omnivla,
  author    = {Noriaki Hirose and Catherine Glossop and Dhruv Shah and Sergey Levine},
  title     = {{OmniVLA}: An Omni-Modal Vision-Language-Action Model for Robot Navigation},
  booktitle = {IEEE International Conference on Robotics and Automation (ICRA)},
  year      = {2026},
  url       = {https://arxiv.org/abs/2509.19480}
}

@article{zhang2025ventura,
  author  = {Arthur Zhang and Xiangyun Meng and Luca Calliari and Dong-Ki Kim and Shayegan Omidshafiei and Joydeep Biswas and Ali Agha and Amirreza Shaban},
  title   = {{VENTURA}: Adapting Image Diffusion Models for Unified Task Conditioned Navigation},
  journal = {arXiv preprint arXiv:2510.01388},
  year    = {2025}
}

@article{ye2026dreamzero,
  author  = {Seonghyeon Ye and Yunhao Ge and Kaiyuan Zheng and Shenyuan Gao and Sihyun Yu and others},
  title   = {World Action Models are Zero-shot Policies},
  journal = {arXiv preprint arXiv:2602.15922},
  year    = {2026}
}

@article{nvidia2026cosmos3,
  author  = {{NVIDIA} and others},
  title   = {{Cosmos 3}: Omnimodal World Models for Physical {AI}},
  journal = {arXiv preprint arXiv:2606.02800},
  year    = {2026}
}

@article{yang2026rise,
  author  = {Jiazhi Yang and Kunyang Lin and Jinwei Li and Wencong Zhang and Tianwei Lin and Longyan Wu and Zhizhong Su and Hao Zhao and Ya-Qin Zhang and Li Chen and Ping Luo and Xiangyu Yue and Hongyang Li},
  title   = {{RISE}: Self-Improving Robot Policy with Compositional World Model},
  journal = {Proceedings of Robotics: Science and Systems},
  year    = {2026}
}

@article{schulman2017ppo,
  author  = {John Schulman and Filip Wolski and Prafulla Dhariwal and Alec Radford and Oleg Klimov},
  title   = {Proximal Policy Optimization Algorithms},
  journal = {arXiv preprint arXiv:1707.06347},
  year    = {2017}
}

@article{shao2024deepseekmath,
  author  = {Zhihong Shao and Peiyi Wang and Qihao Zhu and Runxin Xu and Junxiao Song and Xiao Bi and Haowei Zhang and Mingchuan Zhang and Y. K. Li and Y. Wu and Daya Guo},
  title   = {{DeepSeekMath}: Pushing the Limits of Mathematical Reasoning in Open Language Models},
  journal = {arXiv preprint arXiv:2402.03300},
  year    = {2024}
}

@article{hafner2023dreamerv3,
  author  = {Danijar Hafner and Jurgis Pasukonis and Jimmy Ba and Timothy Lillicrap},
  title   = {Mastering Diverse Control Tasks through World Models},
  journal = {Nature},
  volume  = {640},
  pages   = {647--653},
  year    = {2025},
  doi     = {10.1038/s41586-025-08744-2}
}

@inproceedings{chua2018pets,
  author    = {Kurtland Chua and Roberto Calandra and Rowan McAllister and Sergey Levine},
  title     = {Deep Reinforcement Learning in a Handful of Trials Using Probabilistic Dynamics Models},
  booktitle = {Advances in Neural Information Processing Systems},
  volume    = {31},
  pages     = {4754--4765},
  year      = {2018}
}

@inproceedings{hafner2019planet,
  author    = {Danijar Hafner and Timothy Lillicrap and Ian Fischer and Ruben Villegas and David Ha and Honglak Lee and James Davidson},
  title     = {Learning Latent Dynamics for Planning from Pixels},
  booktitle = {Proceedings of the 36th International Conference on Machine Learning},
  series    = {Proceedings of Machine Learning Research},
  volume    = {97},
  pages     = {2555--2565},
  year      = {2019}
}

@inproceedings{hafner2020dreamer,
  author    = {Danijar Hafner and Timothy Lillicrap and Jimmy Ba and Mohammad Norouzi},
  title     = {Dream to Control: Learning Behaviors by Latent Imagination},
  booktitle = {International Conference on Learning Representations},
  year      = {2020}
}

@inproceedings{yu2020mopo,
  author    = {Tianhe Yu and Garrett Thomas and Lantao Yu and Stefano Ermon and James Zou and Sergey Levine and Chelsea Finn and Tengyu Ma},
  title     = {{MOPO}: Model-Based Offline Policy Optimization},
  booktitle = {Advances in Neural Information Processing Systems},
  volume    = {33},
  pages     = {14129--14142},
  year      = {2020}
}

@inproceedings{kidambi2020morel,
  author    = {Rahul Kidambi and Aravind Rajeswaran and Praneeth Netrapalli and Thorsten Joachims},
  title     = {{MOReL}: Model-Based Offline Reinforcement Learning},
  booktitle = {Advances in Neural Information Processing Systems},
  volume    = {33},
  pages     = {21810--21823},
  year      = {2020}
}

@inproceedings{micheli2023iris,
  author    = {Vincent Micheli and Eloi Alonso and Fran{\c{c}}ois Fleuret},
  title     = {Transformers Are Sample-Efficient World Models},
  booktitle = {International Conference on Learning Representations},
  year      = {2023}
}

@inproceedings{hansen2023modem,
  author    = {Nicklas Hansen and Yixin Lin and Hao Su and Xiaolong Wang and Vikash Kumar and Aravind Rajeswaran},
  title     = {{MoDem}: Accelerating Visual Model-Based Reinforcement Learning with Demonstrations},
  booktitle = {International Conference on Learning Representations},
  year      = {2023}
}

@inproceedings{hansen2024tdmpc2,
  author    = {Nick Hansen and Hao Su and Xiaolong Wang},
  title     = {{TD-MPC2}: Scalable, Robust World Models for Continuous Control},
  booktitle = {International Conference on Learning Representations},
  volume    = {2024},
  pages     = {47376--47405},
  year      = {2024}
}

@inproceedings{janner2019mbpo,
  author    = {Michael Janner and Justin Fu and Marvin Zhang and Sergey Levine},
  title     = {When to Trust Your Model: Model-Based Policy Optimization},
  booktitle = {Advances in Neural Information Processing Systems},
  volume    = {32},
  pages     = {12498--12509},
  year      = {2019}
}

@inproceedings{zhou2025dinowm,
  author    = {Gaoyue Zhou and Hengkai Pan and Yann LeCun and Lerrel Pinto},
  title     = {{DINO-WM}: World Models on Pre-trained Visual Features Enable Zero-shot Planning},
  booktitle = {Proceedings of the 42nd International Conference on Machine Learning},
  series    = {Proceedings of Machine Learning Research},
  volume    = {267},
  pages     = {79115--79135},
  year      = {2025}
}

@inproceedings{wang2023pdml,
  author    = {Xiyao Wang and Wichayaporn Wongkamjan and Ruonan Jia and Furong Huang},
  title     = {Live in the Moment: Learning Dynamics Model Adapted to Evolving Policy},
  booktitle = {Proceedings of the 40th International Conference on Machine Learning},
  series    = {Proceedings of Machine Learning Research},
  volume    = {202},
  pages     = {36470--36493},
  year      = {2023}
}

@inproceedings{sekar2020plan2explore,
  author    = {Ramanan Sekar and Oleh Rybkin and Kostas Daniilidis and Pieter Abbeel and Danijar Hafner and Deepak Pathak},
  title     = {Planning to Explore via Self-Supervised World Models},
  booktitle = {Proceedings of the 37th International Conference on Machine Learning},
  series    = {Proceedings of Machine Learning Research},
  volume    = {119},
  pages     = {8583--8592},
  year      = {2020}
}

@inproceedings{kim2020activewm,
  author    = {Kuno Kim and Megumi Sano and Julian De Freitas and Nick Haber and Daniel Yamins},
  title     = {Active World Model Learning with Progress Curiosity},
  booktitle = {Proceedings of the 37th International Conference on Machine Learning},
  series    = {Proceedings of Machine Learning Research},
  volume    = {119},
  pages     = {5306--5315},
  year      = {2020}
}

@inproceedings{kauvar2023curiousreplay,
  author    = {Isaac Kauvar and Chris Doyle and Linqi Zhou and Nick Haber},
  title     = {Curious Replay for Model-based Adaptation},
  booktitle = {Proceedings of the 40th International Conference on Machine Learning},
  series    = {Proceedings of Machine Learning Research},
  volume    = {202},
  pages     = {16018--16048},
  year      = {2023}
}

@article{anderson2018navigationeval,
  author  = {Peter Anderson and Angel Chang and Devendra Singh Chaplot and Alexey Dosovitskiy and Saurabh Gupta and Vladlen Koltun and Jana Kosecka and Jitendra Malik and Roozbeh Mottaghi and Manolis Savva and Amir R. Zamir},
  title   = {On Evaluation of Embodied Navigation Agents},
  journal = {arXiv preprint arXiv:1807.06757},
  year    = {2018}
}

@article{ilharco2019dtw,
  author  = {Gabriel Ilharco and Vihan Jain and Alexander Ku and Eugene Ie and Jason Baldridge},
  title   = {General Evaluation for Instruction Conditioned Navigation using Dynamic Time Warping},
  journal = {CoRR},
  volume  = {abs/1907.05446},
  year    = {2019},
  url     = {https://arxiv.org/abs/1907.05446}
}

@inproceedings{radford2021clip,
  author    = {Alec Radford and Jong Wook Kim and Chris Hallacy and Aditya Ramesh and Gabriel Goh and Sandhini Agarwal and Girish Sastry and Amanda Askell and Pamela Mishkin and Jack Clark and Gretchen Krueger and Ilya Sutskever},
  title     = {Learning Transferable Visual Models From Natural Language Supervision},
  booktitle = {Proceedings of the 38th International Conference on Machine Learning},
  series    = {Proceedings of Machine Learning Research},
  volume    = {139},
  pages     = {8748--8763},
  year      = {2021}
}

@article{oquab2024dinov2,
  author  = {Maxime Oquab and Timoth{\'e}e Darcet and Th{\'e}o Moutakanni and Huy Vo and Marc Szafraniec and Vasil Khalidov and others},
  title   = {{DINOv2}: Learning Robust Visual Features without Supervision},
  journal = {Transactions on Machine Learning Research},
  year    = {2024}
}

@incollection{schmidhuber2007godel,
  author    = {Juergen Schmidhuber},
  title     = {Goedel Machines: Fully Self-Referential Optimal Universal Self-Improvers},
  booktitle = {Artificial General Intelligence},
  editor    = {Ben Goertzel and Cassio Pennachin},
  publisher = {Springer},
  pages     = {199--226},
  year      = {2007},
  doi       = {10.1007/978-3-540-68677-4_7}
}

@inproceedings{wang2019poet,
  author    = {Rui Wang and Joel Lehman and Jeff Clune and Kenneth O. Stanley},
  title     = {{POET}: Open-Ended Coevolution of Environments and Their Optimized Solutions},
  booktitle = {Proceedings of the Genetic and Evolutionary Computation Conference},
  pages     = {142--151},
  year      = {2019},
  doi       = {10.1145/3321707.3321799}
}

\end{document}